\documentclass[10pt,twocolumn,letterpaper]{article}

\usepackage{iccv}
\usepackage{times}
\usepackage{epsfig}
\usepackage{graphicx}
\usepackage{amsmath}
\usepackage{amssymb}

\usepackage[breaklinks=true,bookmarks=false]{hyperref}

\iccvfinalcopy 

\def\iccvPaperID{****} 

\ificcvfinal\fi

\begin{document}

\title{Dynamic Graph Attention for Referring Expression Comprehension}

\author{Sibei Yang$^{1}$\quad\quad Guanbin Li$^{2}$\thanks{Corresponding author is Guanbin Li. This work was partially supported by the Hong Kong PhD Fellowship, State Key Development Program under Grant No.2016YFB1001004, the National Natural Science Foundation of China under Grant No.61976250 and the Fundamental Research Funds for the Central Universities under Grant No.18lgpy63.} \quad\quad Yizhou Yu$^{1,3}$ \vspace{2mm}\\
$^1$The University of Hong Kong  \quad\quad\quad $^2$Sun Yat-sen University \quad\quad\quad $^3$Deepwise AI Lab\\
{\tt\footnotesize sbyang9@hku.hk}, {\tt\small liguanbin@mail.sysu.edu.cn}, {\tt\small yizhouy@acm.org}
	\vspace{-0mm}
}


\maketitle
\ificcvfinal\thispagestyle{empty}\fi

\begin{abstract}
Referring expression comprehension aims to locate the object instance described by a natural language referring expression in an image. This task is compositional and inherently requires visual reasoning on top of the relationships among the objects in the image. Meanwhile, the visual reasoning process is guided by the linguistic structure of the referring expression. However, existing approaches treat the objects in isolation or only explore the first-order relationships between objects without being aligned with the potential complexity of the expression. Thus it is hard for them to adapt to the grounding of complex referring expressions. In this paper, we explore the problem of referring expression comprehension from the perspective of language-driven visual reasoning, and propose a dynamic graph attention network to perform multi-step reasoning by modeling both the relationships among the objects in the image and the linguistic structure of the expression. In particular, we construct a graph for the image with the nodes and edges corresponding to the objects and their relationships respectively, propose a differential analyzer to predict a language-guided visual reasoning process, and perform stepwise reasoning on top of the graph to update the compound object representation at every node. 
Experimental results demonstrate that the proposed method can not only significantly surpass all existing state-of-the-art algorithms across three common benchmark datasets, but also generate interpretable visual evidences for stepwisely locating the objects referred to in complex language descriptions.
\end{abstract} 

\section{Introduction}
A referring expression is a natural language description of a particular object in an image. Referring expression comprehension thus requires locating the object instance in the image according to a given referring expression. 
It is one of the core tasks in the field of artificial intelligence to realize human-computer communication.

\begin{figure}[t]
\begin{center}
\includegraphics[width=1.0\linewidth]{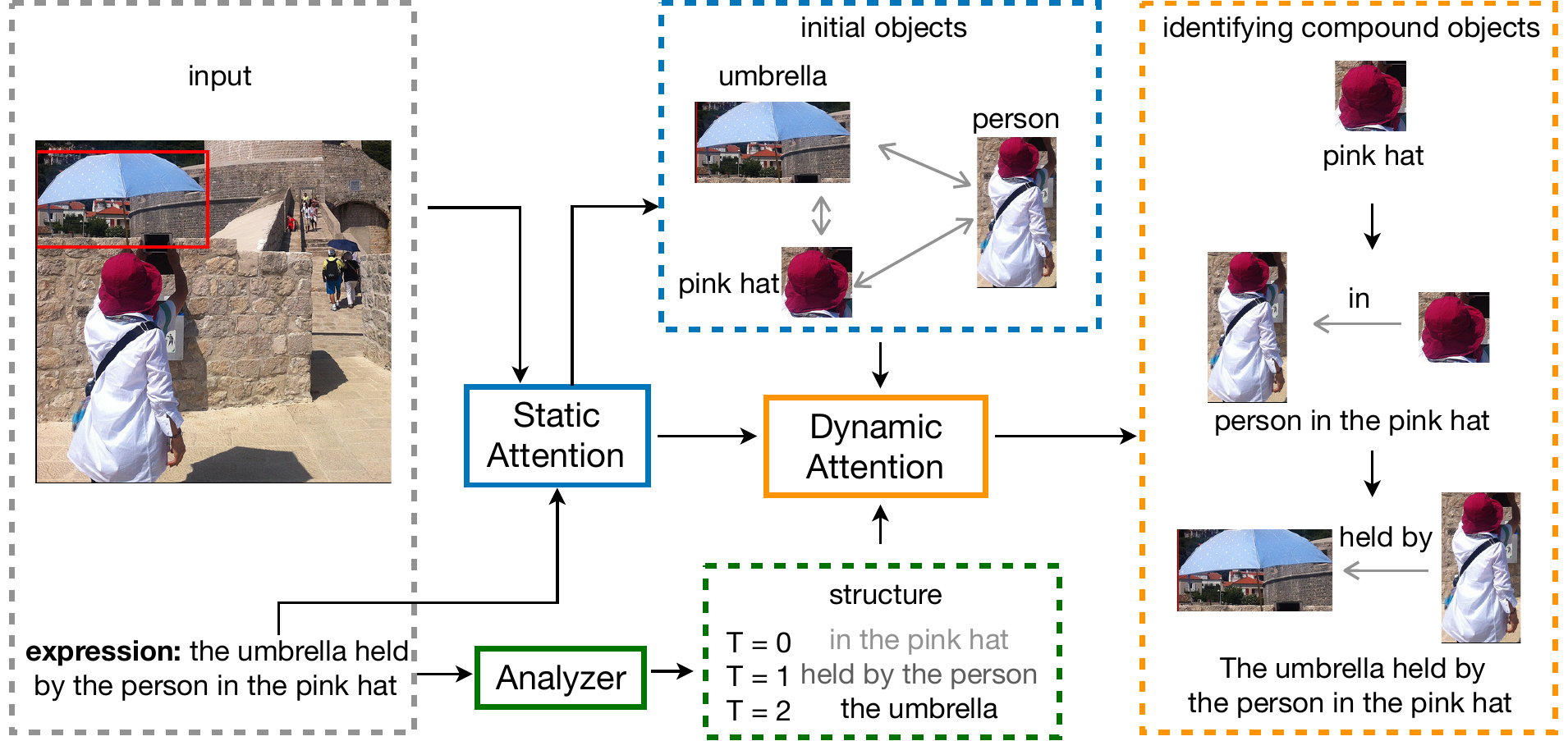}
\end{center}
   \caption{Visual reasoning by Dynamic Graph Attention Network for identifying compound objects. Given an expression and image, the static attention module constructs the multi-modal relation graph; the linguistic structure analyzer prescribes a visual reasoning process based on the expression; the dynamic graph attention module performs visual reasoning on top of the graph by following the prescribed visual reasoning process to identify the compound objects step by step. }
   \vspace{-0.5cm}
\label{fig:intro}
\end{figure}

The core of referring expression comprehension lies in joint understanding of high-level semantics of co-occurring language and visual contents, which inherently involves reasoning. For example, the grounding of the referring expression ``the umbrella held by the person in the pink hat'' requires three-step reasoning~(shown in Figure~\ref{fig:intro}), first locating the pink hat in the image under the guidance of the phrase ``the pink hat'', next identifying the person who is ``in the pink hat'', and finally locating the umbrella which is ``held by'' ``the person in the pink hat''. 
However, almost all the existing approaches for referring expression comprehension do not introduce reasoning or only support single-step reasoning. Meanwhile, the models trained with those approaches have poor interpretability. Among them, the most classic work~\cite{liu2017referring, luo2017comprehension, rohrbach2016grounding, wang2016learning} encodes an expression with an LSTM model \cite{hochreiter1997long}, extracts features of visual objects in the image using CNNs \cite{simonyan2014very, ren2015faster}, and adopts matching loss functions to learn a common feature space for the expression and the visual objects. There also exists work \cite{yu2016modeling, nagaraja2016modeling, wang2019neighbourhood, yang2019cross}, which involves extra pairwise context features or multi-order context features to improve the understanding of the image. However, they generally treat the learning process as a black box without explicit reasoning, and the learned monolithic features do not have adequate competitiveness when complex referring expressions are given. Recently, single-step reasoning~\cite{hu2017modeling, yu2018mattnet} has been proposed by decomposing the expression into different components and matching each component with a corresponding visual region via modular networks. The method in \cite{zhuang2018parallel} is the only one that exploits multi-step reasoning for referring expression comprehension. Its stepwise reasoning is implemented using an LSTM model, which recurrently generates attended visual features while feeding the combination of word embedding and the attended visual features back to the LSTM. However, its stepwise reasoning does not consider the linguistic structure of the expression, and it does not explore the relationships among objects in the image. 

To overcome the aforementioned difficulties, we propose a Dynamic Graph Attention Network (DGA) to achieve a high-level understanding of the expression and the image, and enable the multi-step reasoning of the interactions between the expression and the image. The core ideas behind the proposed DGA come from three aspects, which include expression decomposition based on linguistic structure, object relationships modeling, and multi-step reasoning for identifying compound objects from relations. First, parsing the language structure of the expression is critical because it directly provides the visual reasoning steps for finding the referent. However, it is hard to accurately obtain the linguistic structure of a referring expression as such expressions are usually complex and flexible. Therefore, we resort to a differential analyzer module to predict constituent expressions of the input expression step by step to capture the linguistic structure, and the input expression is represented as a sequence of constituent expressions. Second, it is necessary to take into consideration the relationships among the objects in the image because unambiguous referring expressions normally not only describe the attributes of the referent itself, but also its relationships to other objects in the image \cite{yu2016modeling, hu2017modeling, yang2019cross}. Therefore, the proposed DGA constructs a directed graph over the objects in the image. The nodes and edges of the graph correspond to the objects and relationships among the objects respectively. Last but not the least, 
the DGA performs reasoning over the graph under the guidance of the constituent expressions in a stepwise manner to capture higher-order relationships among the objects and update the compound objects corresponding to each node through graph propagation. 




In summary, this paper has the following contributions:
\begin{itemize}
\setlength{\itemsep}{0pt}
\setlength{\parsep}{0pt}
\setlength{\parskip}{0pt}
    \item It is the first piece of work that explores the problem of referring expression comprehension from the perspective of language-driven visual reasoning in real-world images and expressions. A differential analyzer is proposed to predict a multi-step language-guided visual reasoning process.
    \item A dynamic graph attention network is proposed to perform multi-step visual reasoning on top of a multi-modal relation graph and identify compound objects by following the predicted reasoning process, which is specified as a sequence of constituent expressions.
    \item Experimental results show that the proposed method can not only significantly surpass all existing state-of-the-art algorithms, but also generate visualizable and interpretable results, showing visual evidences for stepwise locating the objects referred to in complex language descriptions.
\end{itemize}
\begin{figure*}[h]
\begin{center}
\includegraphics[width=0.95\linewidth]{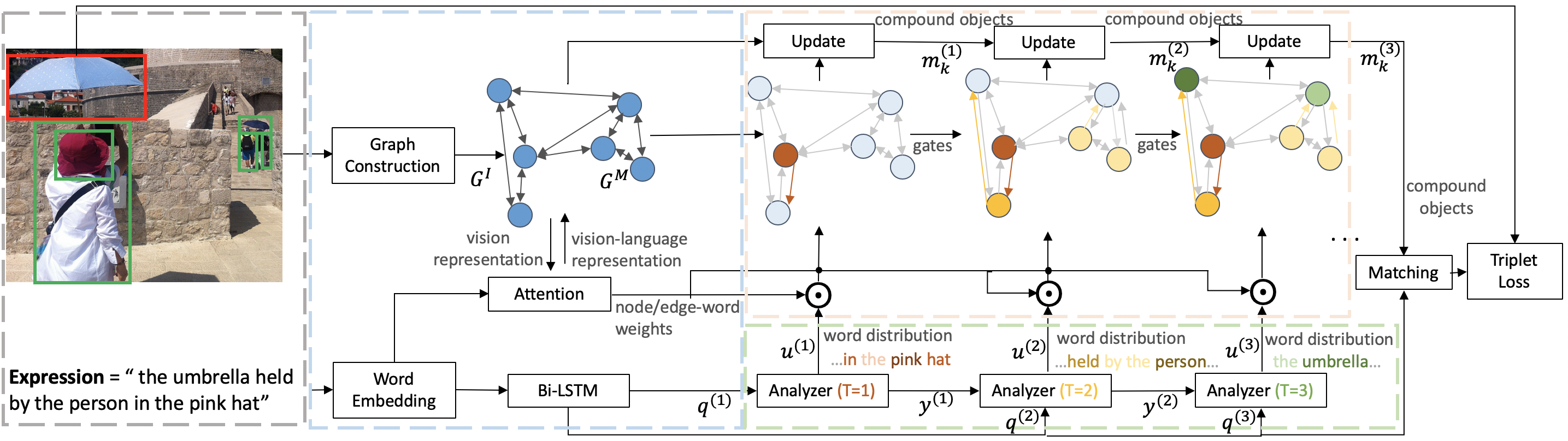}
\end{center}
   \caption{The overall architecture of the Dynamic Graph Attention Network (DGA) for referring expression comprehension. First, the DGA builds a graph over the objects in the image, where the nodes and edges correspond to the objects and relationships respectively, and then fuses the language representation of the expression into the graph; Second, the analyzer learns the language guidance for reasoning by exploring the linguistic structure of the expression. Next, the DGA performs step-wise dynamic reasoning on top of the graph under the guidance of the predicted visual reasoning process which is a sequence of constituent expressions. At each step, the DGA highlights the nodes and edges in the graph by attending the constituent expression over the nodes and edges, and identifies the compound objects for the highlighted nodes by considering their relationships with the compound objects connected by the highlighted edges. Finally, the DGA computes the matching scores between the compound objects and the referring expression. Better view in color, and the different colors represent different steps.}
   \vspace{-0.4cm}
\label{fig:dga}
\end{figure*}

\section{Related Work}
\subsection{Referring Expression Comprehension}
Referring expression comprehension is to locate the object in an image given an input expression. To solve this language-vision multi-modal challenge, it is necessary to learn the correlations between those two modals. Some previous work \cite{luo2017comprehension, rohrbach2016grounding, wang2016learning} independently encodes the inputs in the two modals and learns a common feature space for them. To learn the common feature space, they propose different matching loss functions to optimize, \eg, softmax loss \cite{luo2017comprehension, rohrbach2016grounding} and triplet loss \cite{wang2016learning}. Another work \cite{mao2016generation, yu2016modeling, nagaraja2016modeling} learns to maximize the likelihood of the expression given the referent and the image, and the work inputs the fusion of visual object feature, visual context feature (\eg, entire image CNN feature \cite{mao2016generation}, the visual difference between the objects belonging to the same category in the image \cite{yu2016modeling} and context region CNN features \cite{nagaraja2016modeling}), object location feature and the word embedding to an LSTM to parameterize the distribution. Different from the previous work, recent work \cite{zhuang2018parallel, deng2018visual} adopts co-attention mechanisms to build up the interactions between the expression and the objects in the image.

Those approaches ignore the relationships among objects in the image and the linguistic structure in the expression, which is the key to referring expression comprehension. For an image, they represent the image as a set of independent visual objects \cite{luo2017comprehension, rohrbach2016grounding, wang2016learning, liu2017referring, mao2016generation} or compound objects only including direct relationship \cite{nagaraja2016modeling, yu2016modeling}. For an expression, they encode the expression sequentially and ignore the dependencies in the expression. In order to improve the comprehension, some work \cite{hu2017modeling, yu2018mattnet} designs fixed templates to softly decompose the expression into different semantic components via self-attention, and they compute the language-vision matching scores for each pair of the component and visual region. However, current work is not applicable for expressions that do not conform to the fixed templates. In addition, they ignore the relationships among the visual objects. Recently, \cite{liu2019clevr} explores the visual reasoning for referring expression comprehension in synthetic domain. Different with them, we focus on real-world images and expressions, but do not resort to the guidance of language parsing~(language programs\cite{liu2019clevr}) ground-truth.

To overcome the limitations above, we propose a method to learn to encode the dependencies in the expression and image, and build the interactions between them. We take the linguistic structure into consideration to understand the expression and construct a graph over the visual objects to model the image. And then, their interactions are built up by attention mechanisms.

\subsection{Interpretable Reasoning}
Visual reasoning has drawn much attention because it is essential in the development of Artificial Intelligence. For fulfilling the task of the visual reasoning, the models need to learn reasoning abilities and improve their interpretabilities for the decision rules. There are some existing methods for achieving those requirements. For one-step relational reasoning, the relation networks \cite{santoro2017simple} model pairwise relationships between objects directly. 
For single-step or multi-step reasoning, some work \cite{yang2016stacked, xu2016ask, lu2016hierarchical, hudson2018compositional} explains the reasoning steps by generating updated attention distribution on the image for each step using the attention mechanisms. The other work \cite{andreas2016neural, johnson2017inferring, hu2017learning, cao2018visual} decomposes the reasoning procedure into a sequence of sub-tasks and learns different modular networks to deal with each sub-task.

There are also some methods on referring expression comprehension which attempt to introduce interpretable reasoning. The modular networks are used to improve the interpretabilities of models on referring expression comprehension \cite{hu2017modeling, yu2018mattnet}. \cite{hu2017modeling} decomposes the expression into subject-relationship-object triplets and aligns the textual representations with image regions using localization module or relationship module; however, referring expressions have much richer forms than this fixed subject-relationship-object template. MattNet \cite{yu2018mattnet} decomposes the expressions into three phrases which are corresponding to the subject, location and relationship modules respectively; however, it cannot process multi-step reasoning. The other work \cite{zhuang2018parallel} enables reasoning as a step-wise attention process following the step-wise representation of the expression; however, it treats the expression as the sequence of words, which ignores the linguistic structure of the expression. Different from existing work on referring expression comprehension, we adopt a differential analyzer module to dynamically decompose the expression into its constituent expressions step by step to maintain its linguistic structure and to implement multi-step and dynamic reasoning. 

\section{Dynamic Graph Attention Network}
We introduce a type of network, Dynamic Graph Attention Network (DGA), to address interpretability and multi-step reasoning in referring expression comprehension. Our method performs reasoning by identifying a sequence of compound objects corresponding to partial referring expressions. Our model consists of four main modules: (1) A language-driven differential analyzer (shown inside the green-dotted box in Figure~\ref{fig:dga}), that predicts a visual reasoning process for a referring expression and decomposes the expression into a sequence of constituent expressions, each of which is specified as a soft distribution over the words in the expression. (2) A static graph attention module (shown inside the blue-dotted box in Figure~\ref{fig:dga}), that constructs a directed graph over the visual objects in the image and further builds a multi-modal graph under the guidance of the expression. (3) A dynamic graph attention module (shown inside the orange-dotted box in Figure~\ref{fig:dga}), which enables reasoning on top of the multi-modal graph and identifies compound objects corresponding to constituent expressions. During each reasoning step, the current constituent expression attends the nodes and edges in the graph, and updates the expression-related features of visual objects. 
(4) A matching module, which computes the matching score between an expression and every compound object. 

The overall framework of the proposed DGA is illustrated in Figure~\ref{fig:dga}. In the rest of this section, we elaborate all the modules in this network.
\subsection{Language-Guided Visual Reasoning Process}\label{sec:search}
Referring expressions are complex, and include rich dependencies and nested linguistic structures, which further guide the visual reasoning process.
In theory, natural language parsers can parse grammatical relations among the words in an expression, but existing language parsers are not practical for referring expression comprehension due to highly unrestricted language \cite{yu2018mattnet}. 
Each complex expression is defined by its constituent expressions and the rules used to combine them. We model an expression as a sequence of constituent expressions, and each constituent expression is specified as a soft distribution over the words in the expression.

Given an expression $Q = \{q_l\}_{l=1}^L$ with $L$ words, a DGA network predicts the constituent expression (\ie, a tuple consisting of soft distribution over the words $R^{(t)} = \{r^{(t)}_l\}_{l=1}^L$ and $Q$) corresponding to the compound object at each reasoning step $t$. The DGA's computational process for the distribution is similar to the control unit in \cite{hudson2018compositional}. The DGA first learns an embedding for the words, $\boldsymbol{F} = \{\boldsymbol{f}_l\}_{l=1}^L$, and then encodes the sequence of word embeddings into a vector sequence $\boldsymbol{H} = \{\boldsymbol{h}_l\}_{l=1}^L$ using a bi-directional LSTM \cite{burks1954analysis}, where $\boldsymbol{h}_l$ is the concatenation of the output from the forward and backward LSTMs at the $l$-th word. Meanwhile, the overall expression is represented with a feature vector $\boldsymbol{q}$, which is the concatenation of the last hidden states of both the forward and backward LSTMs. Next, the DGA runs recurrently for $T$ time steps, where $T$ is the number of reasoning steps. During each time step $t$, the DGA transforms the feature vector $\boldsymbol{q}$ into a time-step dependent vector $\boldsymbol{q}^{(t)}$ through a learned linear transform, and concatenates the vector $\boldsymbol{q}^{(t)}$ with the output from the previous time step $\boldsymbol{y}^{(t-1)}$ to form a new vector $\boldsymbol{u}^{(t)}$,
\begin{equation}
\begin{aligned}
  \boldsymbol{q}^{(t)} &= \boldsymbol{W}^{(t)}\boldsymbol{q} +\boldsymbol{b}^{(t)}, \\
  \boldsymbol{u}^{(t)} &= [\boldsymbol{q}^{(t)}; \boldsymbol{y}^{(t-1)}];
\end{aligned}
\end{equation}
where $\boldsymbol{W}^{(t)}$ and $\boldsymbol{b}^{(t)}$ are trainable parameters at time step $t$; $\boldsymbol{y}^{(t-1)}$ is the output at the previous time step $t-1$; $\boldsymbol{u}^{(t)}$ includes the information at previous time steps and the overall information of the expression, and the trainable parameters $\boldsymbol{y}^{(0)}$ is randomly initialized at the beginning of training. Then, the DAG computes the similarity between $\boldsymbol{u}^{(t)}$ and the encoded words $\boldsymbol{H}$ to predict the relevance of each word in visual reasoning during the current time step. The soft distribution over the words at time step $t$, $R^{(t)} = \{r^{(t)}_l\}_{l=1}^L$, is calculated as follows:
\begin{equation}
\begin{aligned}
  \boldsymbol{s}^{(t)} &= \text{relu}(\boldsymbol{W}_{u}\boldsymbol{u}^{(t)} +\boldsymbol{b}_{u}), \\
  a^{(t)}_{l} &= \boldsymbol{W}_{s2}[\text{tanh}(\boldsymbol{W}_{s0}\boldsymbol{s}^{(t)}+\boldsymbol{W}_{s1}\boldsymbol{h}_{l})], \\
  r^{(t)}_{l} &= \frac{\text{exp}(a^{(t)}_{l})}{\sum_{l=1}^L\text{exp}(a^{(t)}_{l})},
\end{aligned}
\end{equation}
where $\boldsymbol{W}_{u}$, $\boldsymbol{b}_{u}$, $\boldsymbol{W}_{s0}$, $\boldsymbol{W}_{s1}$ and $\boldsymbol{W}_{s2}$ are trainable parameters, and they are shared across different time steps. Finally, the output $\boldsymbol{y}^{(t)}$ at time step $t$ is defined as follows:
\begin{equation}
\boldsymbol{y}^{(t)} = \sum_{l=1}^{L}r^{(t)}_l\boldsymbol{h}_l.
\end{equation}
$\boldsymbol{y}^{(t)}$ is part of the input at the next time step $t+1$.

Once we have run this language-guided visual reasoning process for $T$ steps,
the sequence of soft distribution over the words, $\{R^{(t)}\}_{t=1}^T$, can be obtained. The soft constituent expression $(R^{(t)}, Q)$ provides guidance to identify the compound object for time step $t$.

\subsection{Static Graph Attention}
The DGA first constructs a directed graph $G^{I}$ over the visual objects in the image. The nodes of the graph correspond to the visual objects, and the edges correspond to the relationships between objects. Next, the DGA attends the words in the expression over the nodes and edges of the graph $G^{I}$, which builds the connection between the expression and the image, and then sets up a multi-modal graph $G^{M}$. $G^{I}$ models the dependencies among objects in the image while $G^{M}$ enhances $G^{I}$ by representing the interaction between the expression and the image.
\vspace{-0.4cm}
\subsubsection{Graph construction}\label{sec:graph}
\vspace{-0.2cm}
Given an image $I$ with $K$ object proposals $O = \{o_k\}_{k=1}^{K}$ (bounding boxes), the DGA builds a directed graph $G^I = (V, E, \boldsymbol{X}^I)$, where $V = \{v_k\}_{k=1}^K$ is the set of nodes and $v_k$ corresponds to object $o_k$; $E = \{e_{ij}\}_{i,j=1}^K$ is the set of edges and $e_{ij}$ corresponds to the relationship between $o_i$ and $o_j$; $\boldsymbol{X}^I = \{\boldsymbol{x}^I_k\}_{k=1}^K$ is a set of features, and $\boldsymbol{x}^I_k$ is the concatenation of $o_k$'s visual feature $\boldsymbol{x}^o_k$ and $o_k$'s spatial feature $\boldsymbol{p}_k$ ($\boldsymbol{x}^I_k = [\boldsymbol{x}^o_k; \boldsymbol{p}_k]$). In particular, $\boldsymbol{x}^o_k$ is extracted from a pretrained CNN model \cite{simonyan2014very, ren2015faster}, and spatial feature $\boldsymbol{p}_k$ is defined as 
$\boldsymbol{p}_k = \boldsymbol{W}_p[{x_0}_k, {x_1}_k, w_k, h_k, w_kh_k]$, where $({x_0}_k, {x_1}_k)$ are the normalized coordinates of the center of object $o_k$, $w_k$ and $h_k$ are the normalized width and height, and $\boldsymbol{W}_p$ is a trainable parameter.

Similar to \cite{yang2019cross}, we explore the relationship between each pair of object proposals according to their size and locations. For any pair of objects $o_i$ and $o_j$, edge $e_{ij}$ is defined as follows. We compute the relative distance $d_{ij}$, relative angle $\theta_{ij} \in [0, 360)$ (\ie, the angle between the horizontal axis and vector $({x_0}_i - {x_0}_j, {x_1}_i - {x_1}_j)$), and Intersection over Union $m_{ij}$ between them. If $o_i$ includes $o_j$, $e_{ij} = 1$, which means ``inside''; if $o_i$ is covered by $o_j$, $e_{ij} = 2$, which means ``cover''; if none of the above two cases is true and $m_{ij}$ is larger than 0.5, $e_{ij} = 3$, which means ``overlap''; otherwise, when the ratio between $d_{ij}$ and the diagonal length of the image is larger than 0.5, $e_{ij} = 0$, which means ``no relationship''; In the reset of the cases, $e_{ij} = 4 + \lfloor \frac{\theta_{ij} + 22.5}{45} \rfloor$. $e_{ij}=[4,5,...11]$ means ``right'', ``top right'', ``top'', ``top left'', ``left'', ``bottom left'', ``bottom'', and ``bottom right'',  respectively. In summary, $e_{ij} = 0$ means no edge between nodes $v_i$ and $v_j$, and the range of $e_{ij}$ is from $1$ to $N_e = 11$. 

\vspace{-0.2cm}
\subsubsection{Static Attention}\label{sec:static}
\vspace{-0.2cm}
The multi-modal graph $G^{M}$ is defined as $G^{M} = (V, E, \boldsymbol{X}^{M})$, where $V$ and $E$ are as same as the nodes and edges of graph $G^{I}$ respectively, while the features of nodes, $\boldsymbol{X}^{M}$, are computed under the guidance of the expression. Here, we use the word embedding $\boldsymbol{F} = \{\boldsymbol{f}_l\}_{l=1}^L$ mentioned in Section~\ref{sec:search} to represent the expression. 

Words in a referring expression can usually be classified into two types (\ie, entity and relation). We compute the weight of each type, $\boldsymbol{z}_l = [{z_0}_l, {z_1}_l]$, for the $l$-th word represented as $q_l$ as follows,
\begin{equation}
    \begin{aligned}
        {z_0}_l &= \text{sigmoid}(\boldsymbol{W}_{z1}(\boldsymbol{W}_{z0}\boldsymbol{f}_l+\boldsymbol{b}_{z0})+b_{z1}), \\
        {z_1}_l &= 1 - {z_0}_l,
    \end{aligned}
\end{equation}
where $\boldsymbol{W}_{z0}$, $\boldsymbol{W}_{z1}$, $\boldsymbol{b}_{z0}$ and $b_{z1}$ are trainable parameters; ${z_0}_l$ and ${z_1}_l$ are the entity weight and relation weight of word $q_l$ respectively.

Next, we represent the interactions between graph $G^{I} $and the expression by attending the expression over the nodes and edges of the graph. On the basis of the word embedding, $\boldsymbol{F} = \{\boldsymbol{f}_l\}_{l=1}^L$, and the entity weights of words,  $\{{z_0}_l\}_{l=1}^L$, the weighted normalized attention distribution over the nodes of graph $G^{I}$ is defined as follows.
\begin{equation}
    \begin{aligned}
        a_{k,l} &= \boldsymbol{W}_{\alpha2}[\text{tanh}(\boldsymbol{W}_{\alpha1}\boldsymbol{x}_k^{I}+\boldsymbol{W}_{\alpha0}\boldsymbol{f}_l)], \\
        \alpha_{k,l} &= {z_0}_l\frac{\text{exp}(a_{k,l})}{\sum_{k=1}^K\text{exp}(a_{k,l})},
    \end{aligned}
\end{equation}
where $\boldsymbol{W}_{\alpha0}$, $\boldsymbol{W}_{\alpha1}$ and $\boldsymbol{W}_{\alpha2}$ are trainable parameters. $\alpha_{k,l}$ is the weighted normalized attention, indicating the probability of the $l$-th word in the expression referring to node $v_k$. Thus, the language representation $\boldsymbol{c}_k$ at node $v_k$ is computed by aggregating all attention weighted word feature vectors,
\begin{equation}
    \boldsymbol{c}_k = \sum_{l=1}^L\alpha_{k,l}\boldsymbol{f}_l.
\end{equation}

Likewise, we compute a normalized distribution of words over the edges of graph $G^{I}$. Each edge has its own relation type (\ie, 1, ..., 11 as described in Section~\ref{sec:graph}), and the weights for edges are formulated as the weights for edges' types.
\begin{equation}
    \begin{aligned}
    \boldsymbol{\beta}_l = {z_1}_l\text{softmax}(\boldsymbol{W}_{\beta1}\sigma(\boldsymbol{W}_{\beta0}\boldsymbol{f}_l+\boldsymbol{b}_{\beta0})+\boldsymbol{b}_{\beta1}),
    \end{aligned}
\end{equation}
where $\boldsymbol{W}_{\beta0}$, $\boldsymbol{W}_{\beta1}$, $\boldsymbol{b}_{\beta0}$ and $\boldsymbol{b}_{\beta1}$ are trainable parameters; $\sigma$ is the activation function; the softmax function is defined over the $N_e=11$ types; $\beta_{n,l}$ is the $n$-th element of $\boldsymbol{\beta}_l$, which is the weighted probability of the $l$-th word referring to edge type $n \in 1,2,...,N_e$.

Then, we compute the features for the nodes in graph $G^{M}$, $\boldsymbol{X}^{M}$. The feature at node $v_k$, $\boldsymbol{x}_k^{M}$, is a combination of the node feature $\boldsymbol{x}_k^{I}$ of graph $G^{I}$ and the language representation $\boldsymbol{c}_k$,
\begin{equation}
    \begin{aligned}
    \boldsymbol{x}_k^{M} = \boldsymbol{W}_m[\boldsymbol{x}^{I}_k; \boldsymbol{c}_k] + \boldsymbol{b}_m,
    \end{aligned}
\end{equation}
where the $\boldsymbol{W}_m$ and $\boldsymbol{b}_m$ are trainable parameters.

\subsection{Dynamic Graph Attention}
The DGA performs multi-step reasoning on top of the multi-modal graph $G^{M}$ under the guidance of the predicted visual reasoning process $\{R^{(t)}\}_{t=1}^T$ generated from the referring expression (Section~\ref{sec:search}). The DGA's actual reasoning steps takes into account the relationships among the objects in the image as well as the dependencies in the expression. Such reasoning steps start from the initial features $\boldsymbol{X}^M$ at the nodes $V$ of graph $G^{M}$, and these initial features represent individual objects corresponding to the nodes. During the actual reasoning process, the DGA gradually updates the representations of compound objects according to the soft distributions ($\{R^{(t)}\}_{t=1}^T$), the structure of graph $G^{M}$, individual visual objects as well as compound objects in previous time steps.

At each time-step $t$, the DGA maintains a set of memories,  $\boldsymbol{M}^{(t)} = \{\boldsymbol{m}_k^{(t)}\}_{k=1}^K$, to save individual objects ($t=1$) or compound objects ($t>1$) identified in time step $t$, and $\boldsymbol{m}_k^{(t)}$ represents the individual object or compound object corresponding to node $v_k$; meanwhile, it maintains two sets of gates, $P^{(t)} = \{p_k^{(t)}\}_{k=1}^{K}$ and  $\{\nu_{n}^{(t)}\}_{n=1}^{N_e}$, to save the weights of nodes and the weights of edges at the current and all previous time steps. Specifically, $p_k^{(t)}$ represents the weight of node $v_k$ and $\nu_{n}^{(t)}$ represents the weight of edge type $n$. 
Reasoning at time step $t$ is guided by the constituent expression $(R^{(t)} = \{r_l^{(t)}\}_{l=1}^L, Q=\{q_{l}\}_{l=1}^L)$.
By attending the constituent expression over the nodes and edges of graph $G^{M}$, we can obtain the normalized weights of nodes and edges for time step $t$. We compute such weights in two steps. First, we compute the $\gamma_{k,l}^{(t)}$, that represents the probability of the $l$-word referring to node $v_k$, and $\delta_{n,l}^{(t)}$, that represents the probability of the $l$-th word referring to edge type $n$, as weighted the distribution over words, $R^{(t)}$, over the static attention weight, $\alpha_{k,l}$ and $\beta_{n,l}$, introduced in Section~\ref{sec:static},
\begin{equation}
    \begin{aligned}
    \gamma_{k,l}^{(t)} = r^{(t)}_l\alpha_{k,l},\;
    \delta_{n,l}^{(t)} = r^{(t)}_l\beta_{n,l}.
    \end{aligned}
\end{equation}
Second, we compute $\lambda_k^{(t)}$ (or $\mu_n^{(t)}$) that represents the weight of node $v_k$ (or the edge type $n$) being mentioned in time step $t$ as the summation of weights representing individual words in the constituent expression referring to node $v_k$ (or edge type $n$),
\begin{equation}
    \begin{aligned}
    \lambda_k^{(t)} = \sum_{l=1}^{L}\gamma_{k,l}^{(t)},\;
    \mu_n^{(t)} = \sum_{l=1}^{L}\delta_{n,l}^{(t)}.
    \end{aligned}
\end{equation}
Next, we update the gates for every node, $v_k$, and the gates for every type of edge, $n$,
\begin{equation}
    \begin{aligned}
    p^{(t)}_k = \lambda^{(t)}_k + p_k^{(t-1)}, \;
    \nu^{(t)}_n = \mu^{(t)}_n + \nu^{(t-1)}_n.
    \end{aligned}
\end{equation}
Then, we obtain the object feature corresponding to node $v_k$ for time step $t$, $\boldsymbol{m}_k^{(t)}$. When $t = 1$, $\boldsymbol{m}_k^{(t)}$ is set to the feature at node $v_k$ in the multi-modal graph $G^{M}$, $\boldsymbol{x}_k^{M}$. Otherwise, we identify the compound object, $\boldsymbol{m}_{k}$, corresponding to node $v_k$ by considering the nodes connected to $v_k$ as well as compound objects identified in previous time steps,
\begin{equation}
    \begin{aligned}
    \overleftarrow{\boldsymbol{m}}_k^{(t)} &= \sum_{e_{j,k} > 0} {\nu^{(t)}_{e_{j,k}}} (\overleftarrow{\boldsymbol{W}}\boldsymbol{m}_j^{(t-1)} p^{(t-1)}_j + \overleftarrow{\boldsymbol{b}}_{e_{j,k}}), \\
    \widetilde{\boldsymbol{m}}_k^{(t)} &= \widetilde{\boldsymbol{W}}\boldsymbol{m}^{(t-1)}_k + \widetilde{\boldsymbol{b}}, \\
    \boldsymbol{m}^{(t)}_k & = \frac{\lambda^{(t)}_k(\hat{\boldsymbol{W}}(\overleftarrow{\boldsymbol{m}}_k^{(t)} + \widetilde{\boldsymbol{m}}_k^{(t)}) + \hat{\boldsymbol{b}}) + p_k^{(t-1)}\boldsymbol{m}^{(t-1)}_k}{p^{(t)}_k},
    \end{aligned}
\end{equation}
where $\overleftarrow{\boldsymbol{W}}$, $\{\overleftarrow{\boldsymbol{b}}_n\}_{n=1}^{N_e}$, $\widetilde{\boldsymbol{W}}$, $\widetilde{\boldsymbol{b}}$, $\hat{\boldsymbol{W}}$ and $\hat{\boldsymbol{b}}$ are trainable parameters, and they are shared across all time steps. $\overleftarrow{\boldsymbol{m}}_k^{(t)}$ is encoded feature from relationships, $\widetilde{\boldsymbol{m}}_k^{(t)}$ is its updated version, and $\boldsymbol{m}^{(t)}_k$ combines the features from both the current time step and the previous time steps. When $p_k^{(t)}$ is equal to zero, $\boldsymbol{m}^{(t)}_k$ is set to $\boldsymbol{m}^{(t-1)}_k$.

Finally, we use the compound object corresponding to node $v_k$ at the time step $T$ to represent object proposal $o_k$.

\subsection{Matching}
The matching score between proposal $o_k$ and the input expression is defined as follow,
\begin{equation}
    \text{score}_k = \text{L2Norm}(\boldsymbol{W}_{c0}\boldsymbol{m}^{(T)}_k)\odot\text{L2Norm}(\boldsymbol{W}_{c1}\boldsymbol{q}),
\end{equation}
where $\boldsymbol{W}_{c0}$ and $\boldsymbol{W}_{c1}$ are trainable parameters; $\boldsymbol{q}$ is the feature of the entire expression, which is defined in Section~\ref{sec:search}. 

We adopt the triplet loss with online hard negative mining \cite{schroff2015facenet} to train the DGA network. The triplet loss is defined as
\begin{equation}
    \text{loss} = \text{max}(\text{score}_{neg} + \triangle - \text{score}_{gt}, 0),
\end{equation}
where $\text{score}_{neg}$ and $\text{score}_{gt}$ are the matching scores of the negative proposal and the ground-truth proposal respectively; $\triangle$ is the margin. During the inference stage, the proposal with highest matching score is chosen as the prediction.

\begin{table*}
\begin{center}
\resizebox{0.85\textwidth}{!}
{
\begin{tabular}{|l|c|c|c|c|c|c|c|c|c|}
\hline
&  &\multicolumn{3}{|c|}{RefCOCO} & \multicolumn{3}{|c|}{RefCOCO+} & \multicolumn{2}{|c|}{RefCOCOg}  \\ \hline
& feature & val & testA & testB & val & testA & testB & val & test \\ \hline
MMI \cite{mao2016generation} & vgg16 & - & 63.15 & 64.21 & - & 48.73 & 42.13 & - & - \\
Neg Bag \cite{nagaraja2016modeling} & vgg16 & 76.90 & 75.60 & 78.00 & - & - & - & - & 68.40 \\
CG \cite{luo2017comprehension} & vgg16 & - & 74.04 & 73.43 & - & 60.26 & 55.03 & - & - \\
Attr \cite{liu2017referring} & vgg16 & - & 78.85 & 78.07 & - & 61.47 & 57.22 & - & - \\
CMN \cite{hu2017modeling} & vgg16 & - & 75.94 & 79.57 & - & 59.29 & 59.34 & - & - \\
Speaker \cite{yu2016modeling} & vgg16 & 76.18 & 74.39 & 77.30 & 58.94 & 61.29 & 56.24 & - & - \\
Spearker+\textbf{Listener}+Reinforcer\cite{yu2017joint} & vgg16 & 78.36 & 77.97 & 79.86 & 61.33 & 63.10 & 58.19 & 71.32 & 71.72 \\
\textbf{Speaker}+Listener+Reinforcer \cite{yu2017joint} & vgg16 & 79.56 & 78.95 & 80.22 & 62.26 & 64.60 & 59.62 & 71.65 & 71.92  \\
AccumulateAttn \cite{deng2018visual} & vgg16 & 81.27 & 81.17 & 80.01 & 65.56 & 68.76 & 60.63 & - & -  \\
ParallelAttn \cite{zhuang2018parallel} & vgg16 & 81.67 & 80.81 & 81.32 & 64.18 & 66.31 & 61.46 & - & -  \\
MAttNet \cite{yu2018mattnet}  & vgg16 & 80.94 & 79.99 & 82.30 & 63.07 & 65.04 & 61.77 & 73.04 & 72.79  \\
Ours DGA & vgg16 & \textbf{83.73} & \textbf{83.56} & \textbf{82.51} & \textbf{68.99} & \textbf{72.72} & \textbf{62.98} & \textbf{75.76} & \textbf{75.79}  \\ \hline
MAttNet \cite{yu2018mattnet} & resnet101 & 85.65 & 85.26 & 84.57 & 71.01 & 75.13 & 66.17 & 78.10 & 78.12  \\
Ours DGA & resnet101 & \textbf{86.34} & \textbf{86.64} & \textbf{84.79} & \textbf{73.56} & \textbf{78.31} & \textbf{68.15} & \textbf{80.21} & \textbf{80.26}  \\ \hline
\end{tabular}
}
\end{center}
\caption{Comparison with state-of-the-art methods on RefCOCO, RefCOCO+ and RefCOCOg when ground-truth bounding boxes are used. The best performing method is marked in bold.}
\label{tab:ref}
\vspace{-0.3cm}
\end{table*}

\section{Experiments}
\subsection{Datasets}
We have conducted experiments on the following three common benchmark datasets for referring expression comprehension, which were collected from the MSCOCO~\cite{lin2014microsoft} dataset.

\noindent\textbf{RefCOCO}~\cite{yu2016modeling} contains 142,210 referring expressions for 50,000 objects in 19,994 images, which were collected from an interactive game interface~\cite{kazemzadeh2014referitgame}. It is split into train, validation, testA and testB, which has 120,624, 10,834, 5,657 and 5,095 expression-referent pairs, respectively. testA includes images of multiple people while testB includes images with multiple other objects.

\noindent\textbf{RefCOCO+}~\cite{yu2016modeling} has 141,564 expressions for 49,856 objects in 19,992 images collected from an interactive game interface. RefCOCO+ does not contain descriptions of absolute location in the expressions. It is split into train, validation, testA and testB, which has 120,191, 10,758, 5,726 and 4,889 expression-referent pairs, respectively.

\noindent \textbf{RefCOCOg}~\cite{mao2016generation} includes 95,010 long referring expressions for 49,822 objects in 25,799 images collected in a non-interactive setting. RefCOCOg~\cite{nagaraja2016modeling} has 80,512, 4,896 and 9,602 expression-referent pairs for training, validation and testing, respectively.

\subsection{Evaluation and Implementation}
We evaluate the proposed DGA on both ground-truth objects and detected objects. Accuracy is used as the evaluation metric. A prediction is considered correct if the top predicted object is the ground-truth object when ground-truth objects are used, or if the Intersection over Union between the top predicted object and the ground-truth object is larger than 0.5 when detected objects are used.

We follow the similar produce of \cite{yang2019cross} to extract the visual object features of images. Specifically, each object is represented as 2,048-dimensional feature extracted from the pool5 layer of the ResNet-101 based Faster R-CNN model \cite{ren2015faster}. Since some previous methods use VGG-16 \cite{simonyan2014very} as the feature extractor, for the sake of fairness, we also report the results using VGG-16 as backbone. During training, the mini-batch size is set to 64 and we adopt Adam optimizer \cite{kingma2014adam} to update the network parameters. The learning rate is initially set to 0.0005. Margin is set to 0.1 in all our experiments.

\subsection{Comparison with the State of the Art}
We conduct experimental comparison between our proposed DGA and existing state-of-the-art approaches.

\noindent \textbf{Ground-truth objects} Table~\ref{tab:ref} shows quantitative evaluation results on ground-truth objects. Our proposed DGA consistently outperforms existing methods across all the datasets. When the VGG-16 features are used, the DGA improves the average accuracy over the validation and testing sets achieved by the best performing existing approach by 2.00\%, 3.25\% and 2.86\% respectively on the RefCOCO, RefCOCO+ and RefCOCOg datasets. Once we switch to use the ResNet-101 based Faster R-CNN as the backbone, the average accuracy across all the splits is further increased by approximately 4.03\%. These results demonstrate that the linguistic structure of the referring expression and the relationships among the visual objects in the image are conducive to referring expression comprehension.

\noindent \textbf{Detected objects} We have also evaluated the performance of the DGA on automatically detected objects in the three datasets. The detected objects are obtained using Faster R-CNN~\cite{ren2015faster} pretrained on MSCOCO's training images with the images in the validation and testing sets of RefCOCO, RefCOCO+ and RefCOCOg excluded. Since most previous methods report their results using VGG-16 features, for fair comparison, we also adopt VGG-16 features here. The results are shown in Table \ref{tab:det}. The performance drops after we switch from ground-truth objects to detected objects, which is due to detection errors. Nevertheless, the proposed DGA still outperforms all the existing state-of-the-art models, which demonstrates the robustness of the DGA with respect to object detection results.

\begin{table}
\begin{center}
\resizebox{0.48\textwidth}{!}
{
\begin{tabular}{|l|c|c|c|c|c|}
\hline
& \multicolumn{2}{|c|}{RefCOCO} & \multicolumn{2}{|c|}{RefCOCO+} & RefCOCOg  \\ \hline
& testA & testB & testA & testB  & test \\ \hline
MMI \cite{mao2016generation} & 64.90 & 54.51 & 54.03 & 42.81 & - \\
Neg Bag \cite{nagaraja2016modeling} & 58.60 & 56.40 & - & - & 49.50 \\
CG \cite{luo2017comprehension} & 67.94 & 55.18 & 57.05 & 43.33 & - \\
Attr \cite{liu2017referring} & 72.08 & 57.29 & 57.97 & 46.20 & - \\
CMN \cite{hu2017modeling} & 71.03 & 65.77 & 54.32 & 47.76 & - \\
Speaker \cite{yu2016modeling} & 67.64 & 55.16 & 55.81 & 43.43 & - \\
S+\textbf{L}+R \cite{yu2017joint} & 72.94 & 62.98 & 58.68 & 47.68 & 59.63 \\
\textbf{S}+L+R \cite{yu2017joint} & 72.88 & 63.43 & 60.43 & 48.74 & 59.21  \\
ParallelAttn \cite{zhuang2018parallel} & 75.31 & 65.52 & 61.34 & 50.86 & -  \\
Ours DGA & \textbf{78.42} & \textbf{65.53}  & \textbf{69.07} & \textbf{51.99} & \textbf{63.28}  \\ \hline
\end{tabular}
}
\end{center}
\caption{Comparison with the state-of-the-art methods on RefCOCO, RefCOCO+ and RefCOCOg when detected objects are used. The best performing method is marked in bold.}
\label{tab:det}
\vspace{-0.5cm}
\end{table}

\begin{figure*}[h]
\begin{center}
\includegraphics[width=0.91\linewidth]{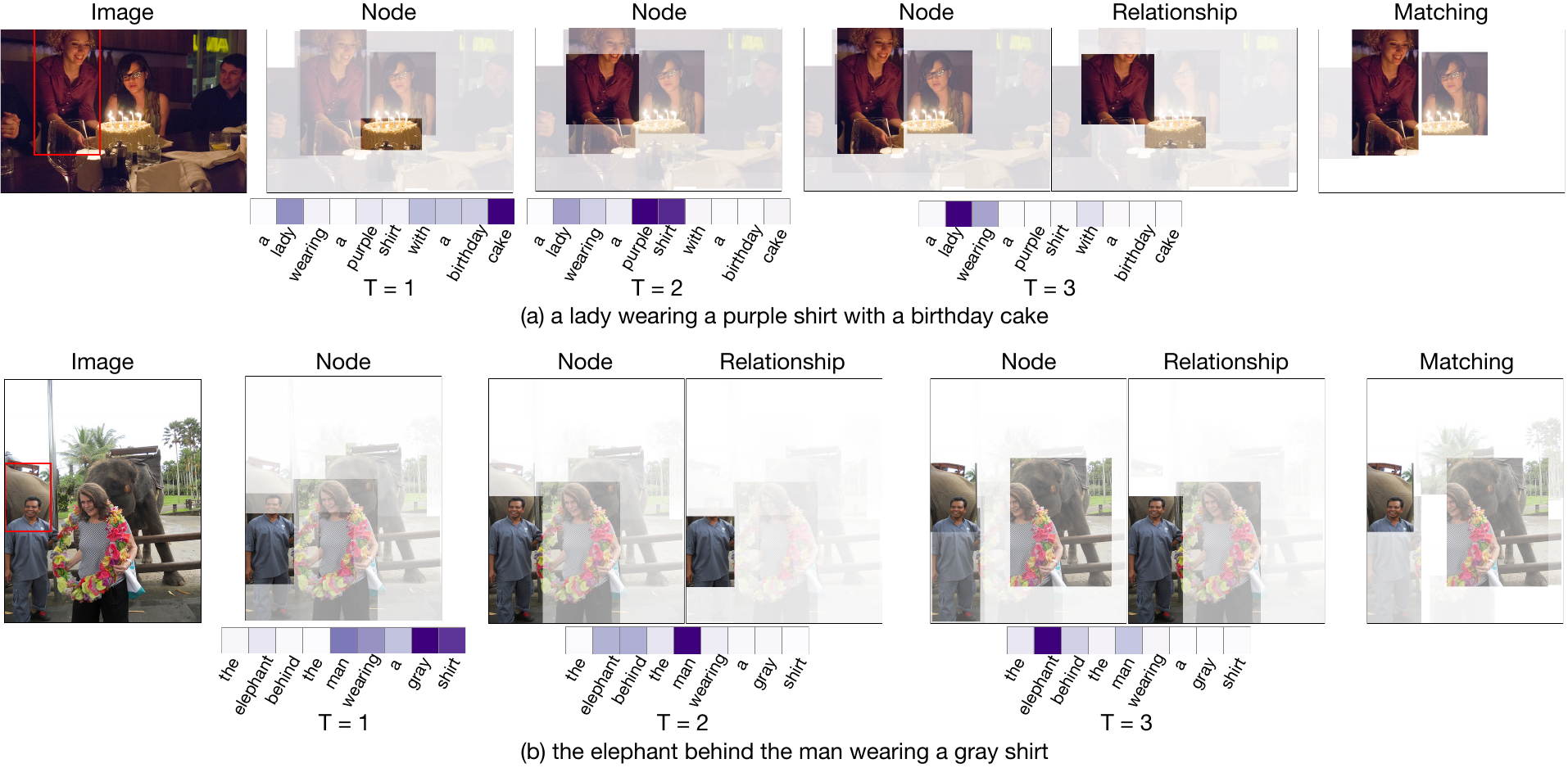}
\end{center}
   \vspace{-0.3cm}
   \caption{Qualitative results showing the iteratively reasoning processes predicted by the DGA, including the word attention weights, node attention maps, relationship attention maps and final matching scores. 
   }
\vspace{-0.3cm}
\label{fig:gre_iccv_vis}
\end{figure*}

\subsection{Qualitative Evaluation}
In order to better explore the reasoning processes learned by the DGA, we study the visualizations of sample results along with their attention distributions produced by the DGA during its iterative computation. At each time step, we visualize the soft distribution over the words to reveal the attended language information during reasoning, and show the attention distribution over graph nodes to indicate the related objects. If a compound object occurs during this time step, we also visualize the relationship distribution by highlighting the other objects that interact with the object that is transformed into the compound object. Moreover, the final matching scores are also provided.

The qualitative evaluation results shown in Figure~\ref{fig:gre_iccv_vis} demonstrates that the proposed DGA can generate visualizable and interpretable evidences for the decision rules. In Figure~\ref{fig:gre_iccv_vis}(a), the expression is parsed into a tree structure, which indicates that the referent ``a lady'' is ``wearing a purple shirt'' and meanwhile it is ``with a birthday cake''. During the first two time steps, the DGA pays more attention to the ``birthday cake'' and the ``purple shirt'' respectively. At the third step, it focuses on the compound object ``a lady wearing a purple shirt with a birthday cake'' by involving the two related objects (\ie ``birthday cake'' and ``purple shirt''). In Figure \ref{fig:gre_iccv_vis}(b), the visual reasoning process forms a chain structure and the DGA gradually identifies the compound objects. At first time step, the DGA attends the ``gray shirt''. Next, it focuses on the compound object ``the man wearing gray shirt'' by connecting ``the man'' with the ``gray shirt''. Then, it shifts focus to the compound object ``the elephant behind the man wearing a gray shirt'' by relating ``the elephant'' to the compound object ``the man wearing gray shirt'' in the last step. The final compound object achieves the highest matching score with the referring expression.

\subsection{Ablation Study}
To demonstrate the effectiveness of the linguistic structure of expressions and multi-step reasoning on top of the relationships among objects in referring expression comprehension, we train four additional models for comparison. The results are shown in Table~\ref{tab:ablation}. The static DGA performs matching between the initial features of nodes in the multi-modal graph with the given referring expression. The performance of the static DGA is worse than the dynamic DGA because the static DGA ignores the relationships among objects and it does not perform reasoning. The DGA with language parser \cite{manning2014stanford} groups the words in the expression into multiple parts, and treats these parts as the constituent expressions to guide reasoning. In comparison to the DGA(3) (a DGA with three time steps), the performance drop of the DGA with language parser demonstrates the crucial role of the proposed analyzer for obtaining the linguistic structure. Next, we explore the number of reasoning steps used in the DGA. The DGA(2) with two steps performs worse than the DGA(3) with three steps and DGA(4) with four steps because DGA(2) only considers direct relationships between objects. The reason why the performance of DGA(3) is better than that of DGA(4) might be that three steps of reasoning are adequate for the datasets used, and any extra steps introduce noise.

\begin{table}
\begin{center}
\resizebox{0.48\textwidth}{!}
{
\begin{tabular}{|l|c|c|c|c|c|c|c|c|}
\hline
&\multicolumn{3}{|c|}{RefCOCO} & \multicolumn{3}{|c|}{RefCOCO+} & \multicolumn{2}{|c|}{RefCOCOg}  \\ \hline
& val & testA & testB & val & testA & testB & val & test \\ \hline
static DGA & 82.10 & 82.13 & 82.08 & 70.56 & 74.71 & 65.31 & 74.45 & 76.52 \\
DGA* & 83.73 & 84.69 & 83.69 & 71.32 & 74.83 & 65.43 & 75.98 & 76.33 \\
DGA(2) & 84.84 & 85.50 & 83.69 & 72.88 & 76.58 & 66.62 & 78.64 & 79.09 \\
DGA(4) & 86.11 & \textbf{86.72} & \textbf{85.65} & 73.34 & 77.10 & 66.95 & 79.17 & 79.90 \\
DGA(3) &\textbf{86.34} & 86.64 & 84.79 & \textbf{73.56} & \textbf{78.31} & \textbf{68.15} & \textbf{80.21} & \textbf{80.26} \\ \hline
\end{tabular}
}
\end{center}
\caption{Ablation study on RefCOCO, RefCOCO+ and RefCOCOg. The number following ``DGA'' indicates the number of reasoning steps used in the model. DGA* means DGA with language parser.}
\label{tab:ablation}
\vspace{-0.5cm}
\end{table}

\section{Conclusion}
In this paper, we have presented Dynamic Graph Attention Networks (DGA) to address referring expression comprehension. A DGA network performs multi-step reasoning on top of the relationships among the objects in an image. This process is guided by the learned linguistic structure of the accompanying referring expression. Experimental results on common benchmark datasets demonstrate that the DGA can not only outperform all existing state-of-the-art methods, but also generate visualizable and interpretable results for the decision rules.

{\small
\bibliographystyle{ieee_fullname}
\bibliography{egbib}
}

\end{document}